\documentclass[runningheads,orivec]{llncs}
\usepackage[T1]{fontenc}
\usepackage{graphicx,verbatim}
\usepackage{booktabs}
\usepackage{cite}
\usepackage{amsmath,amssymb,amsfonts}
\usepackage{algorithmic}
\usepackage{graphicx}
\usepackage{xcolor}
\usepackage{soul}
\usepackage{diagbox}
\usepackage[normalem]{ulem}
\useunder{\uline}{\ul}{}
\usepackage{mathtools}

\usepackage{url}
\usepackage{paralist}
\usepackage{upgreek}
\usepackage{subfigure}
\usepackage[labelfont=bf]{caption}
\usepackage{mathtools}
\usepackage{multirow}
\usepackage{cleveref}
\usepackage{colortbl}
\usepackage{tabularx}
\newcolumntype{Y}{>{\centering\arraybackslash}X}
\definecolor{stdgray}{rgb}{0.300, 0.300, 0.300}
\usepackage[export]{adjustbox} 

\usepackage{tikz} 
\usetikzlibrary{spy}

\newcommand{\etal}{~\textit{et~al.}}

\newif\ifcommentwarn
\commentwarntrue

\def\NVold#1{#1}

\begin{document}
\title{GARD: Gamma-based Anatomical Restoration and Denoising for Retinal OCT
}
\titlerunning{GARD: Gamma-based Anatomical Restoration and Denoising}

\author{
Botond Fazekas\inst{1,2}\orcidID{0000-0001-6158-2612},
Thomas Pinetz\inst{2}\orcidID{0000-0002-6100-2136},
Guilherme Aresta\inst{1,2}\orcidID{0000-0002-4225-2156},
Taha Emre\inst{2}\orcidID{0000-0002-6753-5048},
Hrvoje Bogunovi\'c\inst{1,2}\orcidID{0000-0002-9168-0894}}
%
\authorrunning{Fazekas et al.}
%
\institute{Christian Doppler Laboratory for Artificial Intelligence in Retina, Center for Medical Data Science, Medical University of Vienna, 1090 Vienna, Austria \and
Institute of Artificial Intelligence, Center for Medical Data Science, Medical University of Vienna, 1090 Vienna, Austria}

\maketitle              
\begin{abstract}
Optical Coherence Tomography (OCT) is a vital imaging modality for diagnosing and monitoring retinal diseases. However, OCT images are inherently degraded by speckle noise, which obscures fine details and hinders accurate interpretation.  While numerous denoising methods exist, many struggle to balance noise reduction with the preservation of crucial anatomical structures.  This paper introduces GARD (Gamma-based Anatomical Restoration and Denoising), a novel deep learning approach for OCT image despeckling that leverages the strengths of diffusion probabilistic models.  Unlike conventional diffusion models that assume Gaussian noise, GARD employs a Denoising Diffusion Gamma Model to more accurately reflect the statistical properties of speckle.  Furthermore, we introduce a Noise-Reduced Fidelity Term that utilizes a pre-processed, less-noisy image to guide the denoising process. This crucial addition prevents the reintroduction of high-frequency noise. 
We accelerate the inference process by adapting the Denoising Diffusion Implicit Model framework to our Gamma-based model.  Experiments on a dataset with paired noisy and less-noisy OCT B-scans demonstrate that GARD significantly outperforms traditional denoising methods and state-of-the-art deep learning models in terms of PSNR, SSIM, and MSE. Qualitative results confirm that GARD produces sharper edges and better preserves fine anatomical details.

\keywords{Denoising \and Gamma \and Diffusion models \and OCT \and Retina}
\end{abstract}

\section{Introduction}

Optical Coherence Tomography (OCT) provides cross-sectional images of the retina and other ocular structures \cite{1991_Huang}, allowing the diagnosis and monitoring of a wide range of diseases, from age-related macular degeneration and diabetic retinopathy to glaucoma and optic nerve disorders \cite{1995_Hee}. However,  OCT imaging relies on the interference of backscattered light, making it inherently susceptible to speckle noise. This noise arises from the constructive and destructive interference of coherent light waves scattered from the microscopic structures within the tissue, manifesting as a granular pattern superimposed on the OCT image \cite{1999_Schmitt}.

Speckle noise significantly degrades image quality, obscuring fine details and reducing the contrast between different tissue layers. This can hinder diagnosis, particularly in the early stages of the disease, and can make it difficult to monitor subtle changes over time\cite{2014_Leitgeb}.  Reducing noise levels is therefore highly desirable. One can use repeated acquisitions and average them over time; however, this procedure prolongs acquisition times and introduces registration artifacts in case of patient movements. A reliable OCT denoiser would allow for faster image acquisition times, as fewer scans would be needed to achieve an acceptable signal-to-noise ratio, and potentially allowing similar diagnostic quality with lower-cost, lower-power OCT devices compared to the state-of-the-art. This latter point is especially relevant in developing countries, where access to advanced ophthalmic imaging technology is often limited \cite{2020_WHO}.

Speckle removal (or denoising) in OCT, i.e., the process of suppressing unwanted noise while preserving the essential features, is an active research topic. Early approaches included anisotropic diffusion filtering \cite{2009_Puvanathasan, 2007_Salinas}, non-local means denoising \cite{2015_Aum, 2014_Zhanga}, block matching non-local means denoising (BM3D) \cite{2013_Chong}, wavelet transform-based methods \cite{2017_Zaki, 2015_Kafieh}, and low-rank decomposition-based methods \cite{2016_Kopriva, 2016_Cheng}. Traditional methods have been developed for natural images and therefore have a hard time distinguishing noise from signal in the medical domain, where structures often look vastly different.

Convolutional neural networks (CNNs) are a promising alternative, with, e.g., the conditional General Adversarial Network (cGAN) for OCT despeckling \cite{2018_Ma} outperforming traditional methods. 
However, such supervised approaches usually require a substantial number of noisy/clean pairs for training, which are difficult to acquire in practice.
One prominent approach is self-supervised learning, exemplified by Noise2Void (N2V) \cite{2019_Krull} and its extensions, which eliminates the need for clean target data. A significant advancement in this area is Noise2Void2 (N2V2) \cite{2023_Hoeck}, introducing key modifications to mitigate the checkerboard artifacts often observed with N2V. SCUNet \cite{2023_Zhang}, another state-of-the-art method for various image restoration tasks including denoising, leverages a Swin Transformer-based architecture within a U-Net framework and employs a self-supervised data synthesis technique.

Recently, denoising diffusion probabilistic models (DDPMs) have been successfully applied to medical images\cite{2023_Chung}. DDPMs operate by learning to reverse a process that gradually adds noise to an image, i.e. the denoising process can then be viewed as estimating an intermediate step in this reverse process, given a noisy input. However, standard DDPMs typically assume a Gaussian noise distribution, which is a suboptimal approximation for OCT, where the image is formed by measuring the interference of backscattered light waves.  This interference process leads to speckle noise, which is more accurately modeled by a Gamma distribution \cite{2003_Pircher} due to its ability to capture the positive-only and skewed distribution of intensity fluctuations inherent in coherent imaging \cite{2007_Goodman}.
As an approximation, others have attempted to capture the noise characteristics of OCT images via a DDPM in the logarithmic domain using an entropy based data fidelity term to maintain content consistency\cite{2023_Li}. However, this fidelity term can inadvertently reinforce noise in the input image. In contrast, Nachmani\etal \cite{2021_Nachmani} have already introduced Gamma-based DDPMs (DDGM), demonstrating their effectiveness on synthetic data and natural image inpainting tasks, however, DDGMs have not yet been explored for medical image restoration in particular.

We introduce \textbf{GARD} (\textbf{G}amma-based \textbf{A}natomical \textbf{R}estoration and \textbf{D}enoising) to explore the potential of DDGMs for medical imaging denoising, showcasing its applicability in retinal OCT.
In particular:
\begin{inparaenum}[i)]
    \item \NVold{we adapt and apply a} \textbf{Gamma-distribution based DDPM for OCT}, which more accurately models the statistical properties of speckle noise compared to the commonly used Gaussian assumption,
    \item \NVold{we propose  the \textbf{Noise-Reduced Fidelity Term (NRFT)} that leverages a less noisy image (obtained through techniques like averaging multiple acquisitions) } to guide the denoising process, and 
    \item we evaluate our method on a dataset containing paired sets of OCT images with high and low levels of noise, providing a more \textbf{objective and reliable assessment of denoising performance}.
\end{inparaenum}

\section{Model architecture}

\paragraph{\textbf{Gamma Diffusion Process}}
\NVold{Our approach adapts the diffusion probabilistic model framework \cite{2020_Ho} to address the specific statistical properties of speckle noise in OCT images. The intensity of speckle noise is well-modeled by a Gamma distribution \cite{2003_Pircher}. While other statistical models like the Rayleigh distribution have also been proposed to describe speckle~\cite{2005_Karamata, 2017_Liba}, the Gamma distribution is particularly advantageous for our framework. It not only accurately captures the positive-only and skewed nature of intensity fluctuations inherent in coherent imaging but also possesses mathematical properties that make it highly suitable for the proposed diffusion formulation. Therefore, instead of the standard Gaussian noise used in DDPMs, we employ the Denoising Diffusion Gamma Model (DDGM) process formulated by Nachmani\etal \cite{2021_Nachmani}.}

\NVold{While speckle noise is physically multiplicative in raw, linear-scale OCT data, our model operates on post-processed, display-ready images. These images typically undergo a logarithmic or fourth-square root transformation for dynamic range compression, which renders the noise approximately additive in the domain where our model operates. Consequently, the additive nature of the DDGM forward process is a physically motivated and appropriate choice for OCT denoising.} 

\NVold{The forward noising process is characterized by the following equation:}

\begin{equation}
\mathbf{x}_t = \sqrt{1 - \beta_t} \mathbf{x}_{t-1}  + \left(g_t - \mathbb{E}[g_t] \right)
\label{eq:gamma_forward_step}
\end{equation}
where $\mathbf{x}_t$ is the noisy image at timestep $t$, $\mathbf{x}_{t-1}$ is from the previous timestep. $\beta_t$ is part of a predefined noise schedule. $g_t \sim \Gamma(k_t, \theta_t)$ is a Gamma-distributed random variable with shape parameter $k_t = \frac{\beta_t}{\alpha_t \theta_0^2}$ and scale parameter  $\theta_t = \sqrt{\bar{\alpha}_t}\theta_0$.
Here, $\theta_0$ is a hyperparameter controlling the initial noise level, and $\beta_t$ is part of a predefined noise schedule (similar to DDPMs). This formulation, and the crucial property that the sum of independent Gamma-distributed variables with the same scale is also Gamma-distributed, allows for efficient sampling of the noisy image $\mathbf{x}_t$ at any timestep $t$ directly from the original image $\mathbf{x}_0$.

The reverse process in the DDGM, as derived in \cite{2021_Nachmani}, aims to estimate $x_{t-1}$ from $x_t$.  A neural network, $\epsilon_\theta$, is trained to predict the noise component. The reverse process step is given by:

\begin{equation}
x_{t-1} = \frac{x_t - \frac{1 - \alpha_t}{\sqrt{1 - \bar{\alpha}_t}} \epsilon_{\theta}(x_t, t)}{\sqrt{\bar{\alpha}_t}} + \sigma_t \frac{\bar{g}_t - \mathbb{E}[\bar{g}_t]}{\sqrt{\mathbb{V}[\bar{g}_t]}}
\label{eq:ddgm_reverse}
\end{equation}

\noindent where $\sigma_t$ controls stochasticity, $\mathbb{V}$ is variance and $\bar{g}_t \sim \Gamma(\bar{k}_t, \theta_t)$ with $\bar{k}_t = \sum_{i=1}^t{k_i}$.

To accelerate the reverse process during inference time (denoising), we adapt the Denoising Diffusion Implicit Model (DDIM) sampling approach \cite{2021_Song} to the Gamma diffusion model. Following the DDIM methodology, we achieve deterministic sampling and enable larger denoising steps by setting $\sigma_t = 0$ in Eq.~\ref{eq:ddgm_reverse}. This eliminates the stochastic component of the reverse process, allowing us to skip timesteps during inference.

\paragraph{\textbf{Noise-Reduced Fidelity Term}}
Previous work, e.g. \cite{2023_Li}, has incorporated fidelity terms into the diffusion process to improve data consistency and prevent the generation of spurious artifacts. These terms penalize deviations between the denoised output and the original \emph{noisy} input image.  However, there is a significant drawback, as it can inadvertently reinforce the noise present in the original input, hindering the denoising process. 

\NVold{To overcome this limitation, we propose the \emph{Noise-Reduced Fidelity Term} (NRFT). Our approach intentionally deviates from a traditional MAP estimation framework; instead of enforcing fidelity to the original noisy image, the NRFT leverages a pre-processed, less-noisy estimate to guide the model towards the underlying anatomy. }
The core idea is to use the fidelity term to primarily retain low-frequency information from the original image, while allowing the diffusion model to generate realistic high-frequency details. Instead of comparing the denoised output to the original noisy image $\mathbf{y}_s$, we compare it to a pre-processed version, $\tilde{\mathbf{y}}$, obtained by applying a non-local means (NLM) filter \cite{2005_Buades} to $\mathbf{y}$:

\begin{equation}
    \tilde{\mathbf{y}} = \text{NLM}(\mathbf{y})
    \label{eq:nlm_preprocessing}
\end{equation}

NLM filtering effectively reduces noise while preserving edges and larger structures, making $\tilde{\mathbf{y}}_s$ a more reliable reference for assessing the fidelity of the denoised image.  This pre-processing step provides a cleaner target for the fidelity term, guiding the reverse diffusion process towards a solution that is both consistent with the underlying structure and free from higher-frequency noise. \NVold{The NLM filter was chosen for this pre-processing step due to its practical advantages: it is computationally efficient  and requires no separate training process, unlike deep learning-based alternatives.}

During inference, we incorporate the NRFT by iteratively applying the DDGM reverse process (Eq. \ref{eq:ddgm_reverse}) and then refining the result using the fidelity term.  
Following \cite{2023_Li}, the fidelity term is incorporated by solving the following optimization problem at each reverse step:

\begin{equation}
\tilde{x}^{t-1} \leftarrow \arg\min_{z} \left( z + e^{\tilde{\mathbf{y}} - z} + \mu \left\| z - x^{t-1} \right\|^2 \right)
\label{eq:fidelity_optimization}
\end{equation}

\noindent where $x^{t-1}$ is the output of the DDGM reverse process step (Eq. \ref{eq:ddgm_reverse}), $\tilde{\mathbf{y}}$ is the NLM-filtered image (Eq.~\ref{eq:nlm_preprocessing}), $\tilde{x}^{t-1}$ is the refined estimate at timestep $t-1$ and $\mu$ is a weighting parameter controlling the strength of the fidelity term. We use a Newton's optimization method\cite{1999_Nocedal} to find the solution to this optimization problem. The refined estimate, $\tilde{x}^{t-1}$, then becomes the input, $x^t$, for the next DDGM reverse step. 

\section{Experiments}

\paragraph{\textbf{Datasets}}

Our models were trained in a large and unique dataset with 2\,000 volumes from an investigational High-Res Spectral-Domain OCT device (Heidelberg Engineering, Germany), totaling 86\,819 cross-sectional B-scans.

Quantitative evaluation was done on a prospectively collected dataset with 13 OCT volumes acquired with a commercial Spectralis device (Heidelberg Engineering, Germany), containing pairs of noisy and less-noisy B-scans. Noisy B-scans were obtained from single OCT sweeps, while corresponding less-noisy versions from 30 registered B-scans acquired at the identical anatomical location using the device's \textit{automated real time averaging} (ART). Each volume has 19 B-scans, each with the single-acquisition (noisy) and an averaged, less-noisy version. The perfect registration within each set of B-scans enables a direct pixel-wise comparison, offering a near "gold-standard" for denoising evaluation.

Additionally, a qualitative evaluation dataset was used, comprising Cirrus (Carl Zeiss Meditec, Dublin, CA, USA), Topcon (Topcon Healthcare, Tokyo, Japan), and Spectralis OCT scans from the same patients at the same visit. This dataset was specifically chosen to compare the denoising performance on images from Cirrus and Topcon devices against the less-noisy output of the Spectralis device, which utilizes ART for noise reduction. However, due to differences in acquisition devices and processes, these scans are not registered, limiting their use to visual assessment of denoising results across different OCT platforms.

\paragraph{\textbf{Baselines and ablation studies}}

We compare our proposed model GARD against several baselines. We included a non-deep learning method: Non-Local Means (NLM), and three state-of-the-art image restoration algorithms SCUNet~\cite{2023_Zhang}, Speckle2Speckle~\cite{2022_Goebl} and N2V2~\cite{2023_Hoeck} \NVold{with their publicly available pre-trained weights, to evaluate their off-the-shelf performance}. 
\NVold{In addition to these baselines, we performed an ablation study by testing various diffusion model configurations, including standard Gaussian DDPMs and our Gamma-based DDGMs, both with and without our proposed NRFT and the CDPM fidelity term from~\cite{2023_Li}.}

\paragraph{\textbf{Implementation and Training Details}}

All diffusion models use a U-Net architecture \cite{2015_Ronneberger_CONF}. Both Gaussian and Gamma models used a linear $\beta_t$ schedule from $10^{-4}$ to $0.02$ over $T=1000$ timesteps. Also for Gamma, $\theta_0 = 0.1$ resulted in a noise most similar to typical OCT noise.

The models were trained with the AdamW optimizer (learning rate = $10^{-5}$, batch size = 8) for 500\,000 iterations. Random horizontal flip was used for data augmentation. Models were implemented in PyTorch and trained on one NVIDIA A100 GPU\footnote[1]{The source code of our project is available at \url{https://github.com/ABotond/GARD}.}. For non-local-mean filtering we used the fast implementation from the scikit-image package. In NRFT $\mu$ was set to 10 to ensure high consistency with the input image. Instead of performing the full reverse process from $t=T$ to $t=0$, we started the denoising process at $t=70$ and sampled every 10th timestep. This significantly reduces the number of inference steps required without sacrificing image quality.

\paragraph{\textbf{Evaluation Metrics}}
We quantitatively assessed denoising performance using three standard metrics: Peak Signal-to-Noise Ratio (PSNR), Structural Similarity Index (SSIM), and Mean Square Error (MSE).

In addition, we performed Wilcoxon signed-rank tests to determine the statistical significance of the observed differences in performance between GARD and the baselines.

\section{Results and discussion}

GARD achieves the best performance in all metrics (Table~\ref{tab:results}), outperforming traditional methods (NLM), state-of-the-art deep learning methods (SCUNet, N2V2), \NVold{specialized self-supervised speckle denoisers (Speckle2Speckle),} and other diffusion-based models (DDPM, CPDM, and their variants). Notably, our method achieves a PSNR improvement of 0.31 dB over SCUNet, 0.34 dB over the standard DDPM and 0.23 dB over the vanilla DDGM. This demonstrates the combined effectiveness of the Gamma diffusion process and our proposed Noise-Reduced Fidelity Term (NRFT) for OCT image despeckling. GARD is significantly better for all metrics and methods with $p<0.01$.

The ablation studies, comparing DDGM, DDGM+CPDM and GARD highlight the contribution of the NRFT. The vanilla DDGM already shows competitive performance, ranking second best across all metrics. The fidelity term of \cite{2023_Li} (CPDM) negatively impacts performance because it enforces consistency with the noisy input, effectively reintroducing the high-frequency noise that the diffusion process aims to eliminate. In contrast, our NRFT significantly improves performance over the vanilla DDGM, demonstrating its benefit. Interestingly, while the NRFT significantly boosts the performance of the DDGM, it results in a minor performance degradation for the standard DDPM (Table~\ref{tab:results}). This suggests a potential synergistic effect between our proposed fidelity term and the Gamma-based diffusion process. The NLM-filtered image, by preserving anatomical edges, appears to provide a more effective structural guide for the Gamma model which better aligns with the underlying speckle statistics. Conversely, for the mismatched Gaussian assumption in the standard DDPM, this specific form of guidance may be less compatible, failing to yield a similar improvement. These results validate our hypothesis about using a noise-reduced image for fidelity.

\begin{table}[!thb]
\centering
\caption{Performance on the paired dataset.  Mean (standard-deviation) shown for SSIM, PSNR (dB), and MSE. Bold indicates the best performance, underlined indicates the second best. The different configurations are compared against their less-noisier counterparts with higher ART number. \mbox{*} denotes statistically significant difference from GARD with $p < 0.01$.} \label{tab:results}
\begin{tabularx}{\textwidth}{lYYY} 
    \toprule
    \textbf{Model}             & \textbf{SSIM} $\uparrow$                                 & \textbf{PSNR} $\uparrow$                         & \textbf{MSE} $\downarrow$                          \\ 
    \midrule
    \rowcolor[rgb]{0.92,0.92,0.92} Noisy input                 & $0.43^*$ \scriptsize{\textcolor{stdgray}{(0.04)}}                        & $24.80^*$ \scriptsize{\textcolor{stdgray}{(1.15)}}                        & $222.52^*$ \scriptsize{\textcolor{stdgray}{(56.86)}}                         \\
    NLM \cite{2005_Buades}      & $0.52^*$ \scriptsize{\textcolor{stdgray}{(0.10)}}                        & $26.96^*$ \scriptsize{\textcolor{stdgray}{(1.95)}}                        & $144.78^*$ \scriptsize{\textcolor{stdgray}{(67.28)}}                         \\
    \rowcolor[rgb]{0.92,0.92,0.92}Speckle2Speckle \cite{2022_Goebl}      & $0.49^*$ \scriptsize{\textcolor{stdgray}{(0.07)}}                        & $26.59^*$ \scriptsize{\textcolor{stdgray}{(1.06)}}                        & $147.08^*$ \scriptsize{\textcolor{stdgray}{(38.13)}}                         \\
    N2V2 \cite{2023_Hoeck}      & $0.31^*$ \scriptsize{\textcolor{stdgray}{(0.06)}}                        & $24.76^*$ \scriptsize{\textcolor{stdgray}{(1.26)}}                        & $226.69^*$ \scriptsize{\textcolor{stdgray}{(66.75)}}                         \\
    \rowcolor[rgb]{0.92,0.92,0.92}SCUNet \cite{2023_Zhang}        & $0.55^*$ \scriptsize{\textcolor{stdgray}{(0.08)}}                     & $28.10^*$ \scriptsize{\textcolor{stdgray}{(1.55)}}                     & $107.89^*$ \scriptsize{\textcolor{stdgray}{(45.93)}}                      \\
    DDPM\cite{2023_Chung}         & $0.54^*$ \scriptsize{\textcolor{stdgray}{(0.08)}}        & $27.85^*$ \scriptsize{\textcolor{stdgray}{(1.38)}}        & $112.04^*$ \scriptsize{\textcolor{stdgray}{(35.28)}}         \\
    \rowcolor[rgb]{0.92,0.92,0.92}DDPM\cite{2023_Chung} + CPDM & $0.51^*$ \scriptsize{\textcolor{stdgray}{(0.07)}}       & $27.21^*$ \scriptsize{\textcolor{stdgray}{(1.51)}}       & $130.78^*$ \scriptsize{\textcolor{stdgray}{(42.56)}}        \\
    DDPM\cite{2023_Chung} + NRFT & $0.53^*$ \scriptsize{\textcolor{stdgray}{(0.08)}}       & $27.62^*$ \scriptsize{\textcolor{stdgray}{(1.54)}}       & $119.52^*$ \scriptsize{\textcolor{stdgray}{(42.58)}}        \\
    \rowcolor[rgb]{0.92,0.92,0.92}CPDM \cite{2023_Li}       & $0.44^*$ \scriptsize{\textcolor{stdgray}{(0.04)}} & $25.39^*$ \scriptsize{\textcolor{stdgray}{(1.12)}} & $194.03^*$ \scriptsize{\textcolor{stdgray}{(48.72)}}  \\
    CPDM \cite{2023_Li} + NRFT & $0.47^*$ \scriptsize{\textcolor{stdgray}{(0.07)}} & $26.10^*$ \scriptsize{\textcolor{stdgray}{(1.59)}} & $170.07^*$ \scriptsize{\textcolor{stdgray}{(61.16)}}  \\ 
    \midrule
    DDGM                & $\underline{0.56}^*$ \scriptsize{\textcolor{stdgray}{(0.08)}}     & $\underline{28.16}^*$ \scriptsize{\textcolor{stdgray}{(1.46)}}     & $\underline{105.39}^*$ \scriptsize{\textcolor{stdgray}{(39.92)}}      \\
    \rowcolor[rgb]{0.92,0.92,0.92}DDGM + CPDM & $0.53^*$ \scriptsize{\textcolor{stdgray}{(0.08)}}          & $27.69^*$ \scriptsize{\textcolor{stdgray}{(1.55)}}          & $118.24^*$ \scriptsize{\textcolor{stdgray}{(45.76)}}           \\
    \textbf{GARD} & $\mathbf{0.58}$ \scriptsize{\textcolor{stdgray}{(0.09)}}                & $\mathbf{28.25}$ \scriptsize{\textcolor{stdgray}{(1.54)}}                & $\mathbf{103.95}$ \scriptsize{\textcolor{stdgray}{(41.91)}}                 \\
    \bottomrule
\end{tabularx}
\end{table}

\begin{figure*}[!thb]
    \centering
    \setlength{\tabcolsep}{1pt}
    
\resizebox{\linewidth}{!}{%
    \begin{tabular}{cccccc}
            \multicolumn{1}{c}{\small{Original}} & 
            \multicolumn{1}{c}{\small{Crop}} & \multicolumn{1}{c}{\small{SCUNet}} & \multicolumn{1}{c}{\small{DDPM}} & \multicolumn{1}{c}{\small{GARD}} & \multicolumn{1}{c}{\small{Target}} \\
        \includegraphics[width=0.167\textwidth]{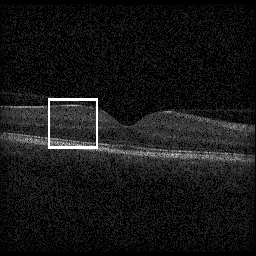} &
        \includegraphics[width=0.167\textwidth]{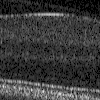} &
        \includegraphics[width=0.167\textwidth]{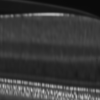} &
        \includegraphics[width=0.167\textwidth]{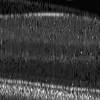} &
        \includegraphics[width=0.167\textwidth]{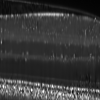} &
        \includegraphics[width=0.167\textwidth]{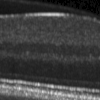} \\
              \includegraphics[width=0.167\textwidth]{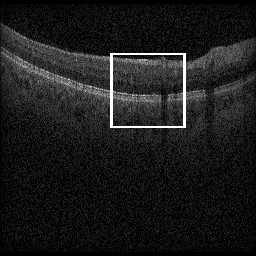} &
        \includegraphics[width=0.167\textwidth]{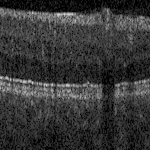} &
        \includegraphics[width=0.167\textwidth]{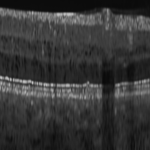} &
        \includegraphics[width=0.167\textwidth]{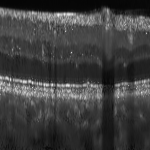} &
        \includegraphics[width=0.167\textwidth]{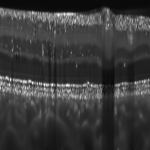} &
        \includegraphics[width=0.167\textwidth]{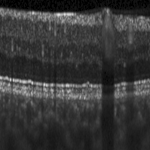} \\
                \includegraphics[width=0.167\textwidth]{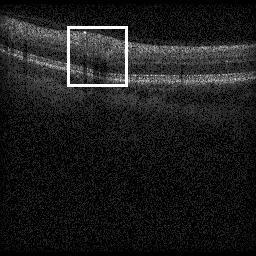} &
        \includegraphics[width=0.167\textwidth]{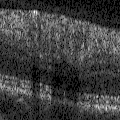} &
        \includegraphics[width=0.167\textwidth]{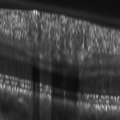} &
        \includegraphics[width=0.167\textwidth]{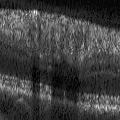} &
        \includegraphics[width=0.167\textwidth]{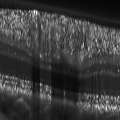} &
        \includegraphics[width=0.167\textwidth]{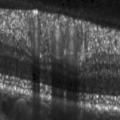} \\
                \includegraphics[width=0.167\textwidth]{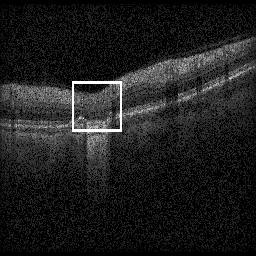} &
        \includegraphics[width=0.167\textwidth]{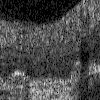} &
        \includegraphics[width=0.167\textwidth]{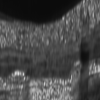} &
        \includegraphics[width=0.167\textwidth]{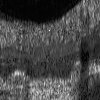} &
        \includegraphics[width=0.167\textwidth]{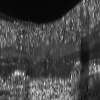} &
        \includegraphics[width=0.167\textwidth]{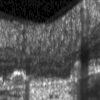} \\
    \end{tabular}
    }
    \caption{Qualitative denoising results on a OCT B-scan. Columns: Original noisy image with crop region indicated, Cropped noisy image, and the results for the best performing baseline model SCUNet,  DDPM,  GARD (Ours), and  less-noisy reference target. GARD provides the sharpest edges and best detail preservation.}
    \label{fig:qual_results}
\end{figure*}

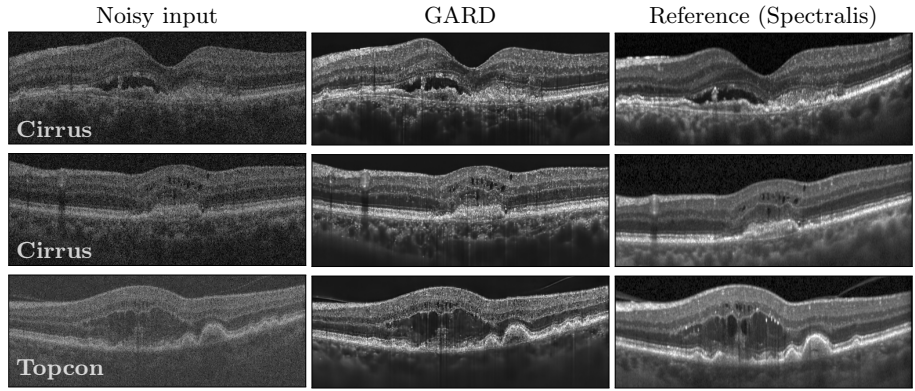
\begin{figure*}[!bht]
    \centering
    \setlength{\tabcolsep}{1pt}
    
\resizebox{\linewidth}{!}{%
    \begin{tabular}{ccc}
            \multicolumn{1}{c}{\small{Noisy input}} & 
            \multicolumn{1}{c}{\small{GARD}} &  \multicolumn{1}{c}{\small{Reference (Spectralis)}} \\
        \begin{tikzpicture}
    \node[inner sep=0] (image) at (0,0) {\includegraphics[width=0.33\textwidth, frame, clip, trim=0px 160px 0px 160px]{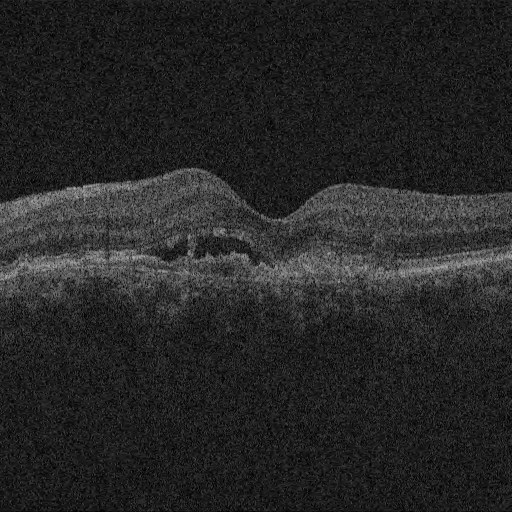}};
    \node[anchor=south west, text=white, opacity=0.8] at ([xshift=0pt,yshift=0pt]image.south west) {\textbf{Cirrus}};
 \end{tikzpicture} &

        \includegraphics[width=0.33\textwidth, frame, clip, trim=0px 160px 0px 160px]{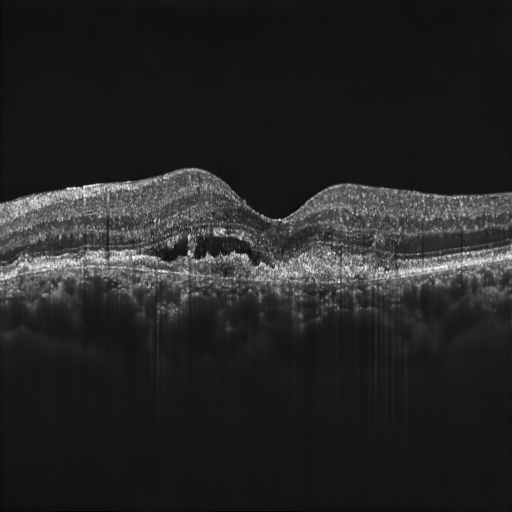} &
        \includegraphics[width=0.33\textwidth, frame, clip, trim=0px 210px 0px 110px]{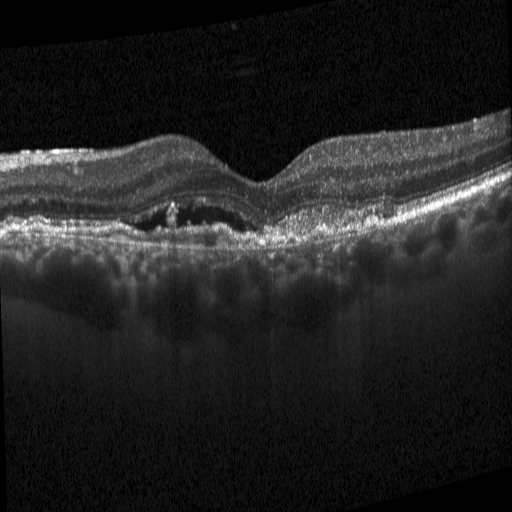} \\
        \begin{tikzpicture}
    \node[inner sep=0] (image) at (0,0) {        \includegraphics[width=0.33\textwidth, frame, clip, trim=0px 200px 0px 120px]{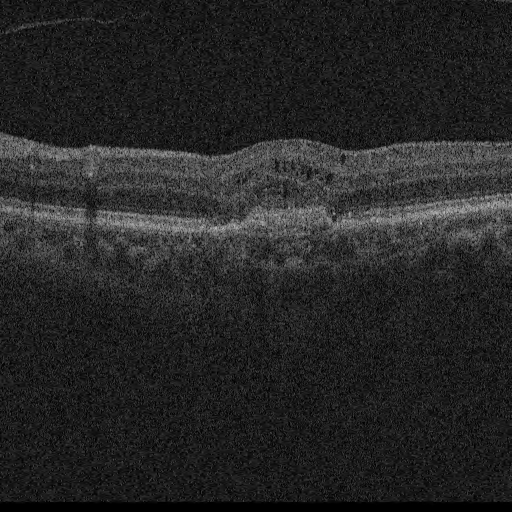}};
    \node[anchor=south west, text=white, opacity=0.8] at ([xshift=0pt,yshift=0pt]image.south west) {\textbf{Cirrus}};
 \end{tikzpicture} &
        \includegraphics[width=0.33\textwidth, frame, clip, trim=0px 200px 0px 120px]{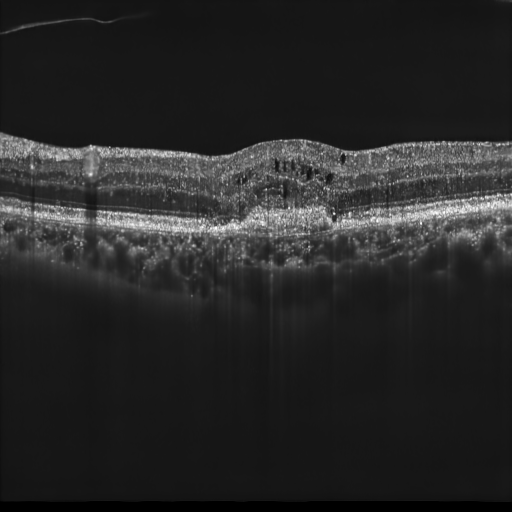} &
        \includegraphics[width=0.33\textwidth, frame, clip, trim=0px 210px 0px 110px]{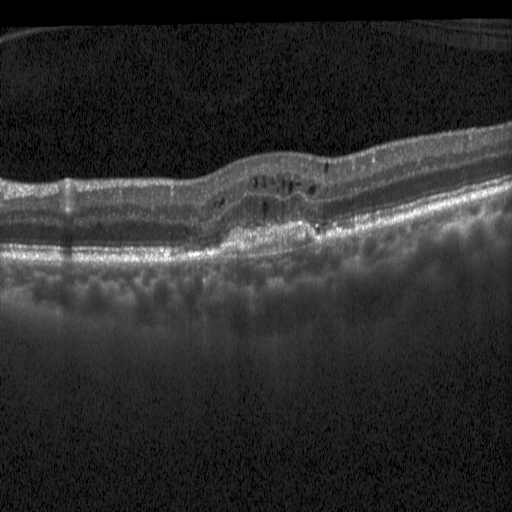} \\
                \begin{tikzpicture}
    \node[inner sep=0] (image) at (0,0) {        \includegraphics[width=0.33\textwidth, frame, clip, trim=0px 240px 0px 80px]{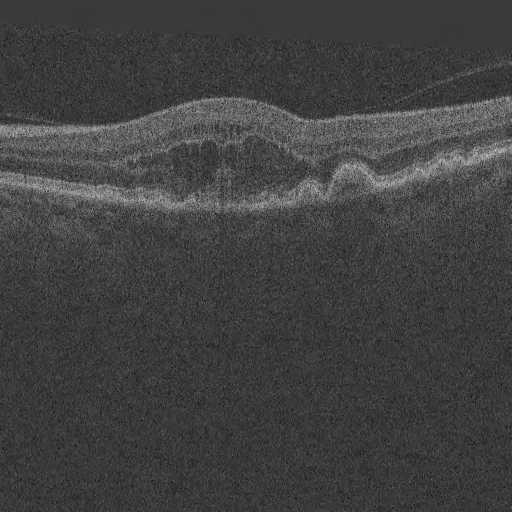}};
    \node[anchor=south west, text=white, opacity=0.8] at ([xshift=0pt,yshift=0pt]image.south west) {\textbf{Topcon}};
 \end{tikzpicture} &
        \includegraphics[width=0.33\textwidth, frame, clip, trim=0px 240px 0px 80px]{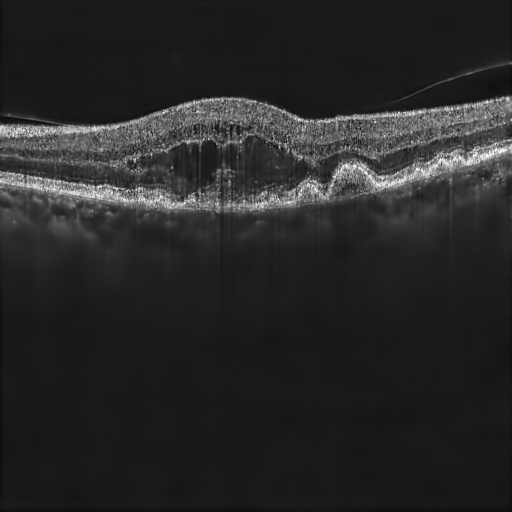} &
        \includegraphics[width=0.33\textwidth, frame, clip, trim=0px 150px 0px 170px]{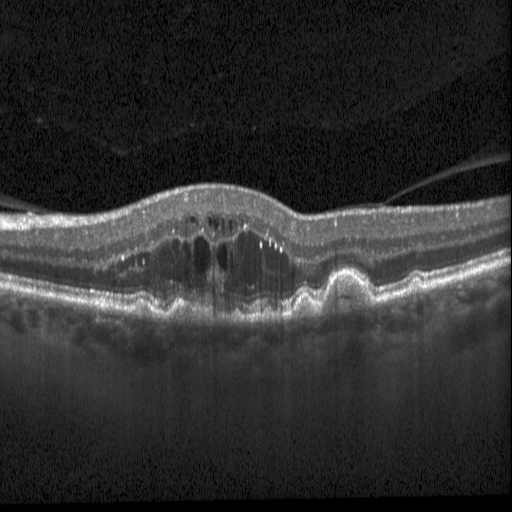} \\
    \end{tabular}
    }
    \caption{Qualitative comparison of GARD denoising on OCT scans from different devices. Each row represents a different eye. Spectralis scans were acquired with ART and serve as less-noisy references.}
    \label{fig:qual_results_multidevice}
\end{figure*}

Visual inspection (Fig.~\ref{fig:qual_results}) supports the quantitative findings. The noisy input exhibits substantial speckle. SCUNet reduces noise but still leaves some fine structures missing, especially small regions with high reflectivity.  DDPM shows considerable improvement, but some fine structures are still missing.  GARD delivers the visually best result, with sharper edges and better preservation of subtle anatomical details, confirming the potential for improved diagnostic utility. We visually evaluated our method on scans from different device vendors (Fig.~\ref{fig:qual_results_multidevice}), and it considerably improves their quality. However, due to the lack of appropriate metrics, we present these results only for illustrative purposes.

While GARD shows superior performance on our paired dataset and promising qualitative results on scans from different vendors, we acknowledge that the quantitative evaluation was conducted on data from a single device type. This is primarily due to the difficulty in acquiring perfectly registered noisy and less-noisy paired scans from multiple commercial devices. Therefore, extensive validation on a wider range of devices and clinical conditions remains a key direction for future work.

\section{Conclusion}
We introduced GARD, a denoising diffusion gamma model (DDGM) with a noise-reduced fidelity term (NRFT) for speckle reduction in retinal OCT images. Our approach, leveraging the Gamma distribution for accurate noise modeling, demonstrated superior quantitative and qualitative performance compared to traditional and deep learning baselines, including other diffusion model variants. The NRFT, utilizing a pre-processed less-noisy image, proved crucial for avoiding noise reinforcement, leading to sharper edges and better fine detail preservation. This improved image quality has the potential to enhance diagnostic accuracy of retinal diseases, especially in underserved regions where lower-cost OCT devices, enhanced by our denoising method, could provide clinically useful images.
While promising, our quantitative evaluation was performed on a single, specialized dataset due to the rarity of paired noisy and less-noisy OCT data. Future work will focus on validating our method on diverse datasets and exploring its applicability to other imaging modalities affected by speckle noise.

\section*{Acknowledgements}
The financial support by the the Christian Doppler Research Association, Austrian Federal Ministry of Economy, Energy and Tourism, the National Foundation for Research, Technology and Development, and Heidelberg Engineering is gratefully acknowledged.

%
%
%

\bibliographystyle{splncs04}
\bibliography{paper8}

\end{document}